\documentclass{article}

\usepackage{arxiv}

\usepackage[utf8]{inputenc} 
\usepackage[T1]{fontenc}    
\usepackage{hyperref}       
\usepackage{url}            
\usepackage{booktabs}       
\usepackage{amsfonts}       
\usepackage{nicefrac}       
\usepackage{microtype}      
\usepackage{lipsum}
\usepackage{graphicx}
\graphicspath{ {./images/} }

\title{Improving Deep Learning Model Robustness Against Adversarial Attack by Increasing the Network Capacity}

\author{
 Marco Marchetti\\
  Department of Computer and Information Sciences\\
  Northumbria University\\
  Newcastle upon Tyne, NE1 8ST \\
  \texttt{m.marchetti@northumbria.ac.uk} \\
   \And
 Edmond S. L. Ho\thanks{corresponding author}\\
  Department of Computer and Information Sciences\\
  Northumbria University\\
  Newcastle upon Tyne, NE1 8ST \\
  \texttt{e.ho@northumbria.ac.uk} \\
}

\begin{document}
\maketitle
\begin{abstract}
Nowadays, we are more and more reliant on Deep Learning (DL) models and thus it is essential to safeguard the security of these systems. This paper explores the security issues in Deep Learning and analyses, through the use of experiments, the way forward to build more resilient models. Experiments are conducted to identify the strengths and weaknesses of a new approach to improve the robustness of DL models against adversarial attacks. The results show improvements and new ideas that can be used as recommendations for researchers and practitioners to create increasingly better DL algorithms.
\end{abstract}


\section{Introduction}
Security in Deep Learning (DL) is becoming increasingly important due to the extensive presence of its application in our daily life \cite{Ho:2022}. 
DL is also bringing benefits in a lot of aspects of the computing field, such as solving complex problems efficiently, great understanding of unstructured data, reduction of costs and high-quality results, etc. However, DL still faces great disadvantages, such as the need for a huge amount of data and the computational intensity \cite{Tariq2020} constitute important cons for the use of Deep Learning.

This implies DL applications are having an increasing demand for resources, therefore outsourcing techniques are required, such as MLaaS (Machine Learning as a Service). Here, data is sent to the cloud and the DL algorithm is there executed, freeing the users from running heavy programs on their machines. However, this outsourcing of data creates data privacy and security concerns because there is a DL algorithm exposure to adversarial attacks \cite{Xu:2019}.


To this regard, Xue et al. \cite{Xue:Access2020} clearly sum up the reasons why DL can be attacked. This includes: the outsourcing of the training phase, the integration into the network of third parties pre-trained models and the lack of data validations for data coming from untrusted users or third parties. Despite the fact that these working paradigms bring undoubtedly advantages, it might also rise security issues.

Essentially, adversarial attacks mainly target MLaaS or DML (Distributed Machine Learning) systems \cite{Chen:Access2020}, for their nature of relying on the Internet. In particular, in a distributed situation the thing that, by expanding, generates more probable threats is the attack surface, i.e. the sum of the different points where an attacker can try to enter data or extract data from a software environment \cite{Manadhata:TSE:2011}. Additionally, it needs to be pointed out that even the most famous Deep Learning frameworks, such as TensorFlow, Caffe and Torch, present vulnerabilities. Xiao et al. \cite{Xiao:SPW2018} have studied the risks caused by those and their impact on common DL applications, particularly voice recognition and image classification. To tackle the problem of guaranteeing security while outsourcing potentially sensitive data, Ghodsi and Garg 
have built 
SafetyNet \cite{Ghodsi:NIPS:2017} which implements a framework to enable an untrusted server (the cloud in our case) to provide a short mathematical proof of the correctness of the tasks being performed for the client (MLaaS). By this, the DNN prevents any poisoning attacks by detecting them. 


Adversarial attacks can be of two kinds \cite{Qayyum:2020}: a poisoning attack or an evasion attack. The first type of attack affects the training phase of the learning process by manipulating the training data \cite{Biggio:ICML2012} with the use of the so-called poisoning examples. Whereas, an adversarial attack on the interface phase (or predictions, see Appendix A) of the learning process is called evasion attack, where an attacker manipulates the test data or the real-time inputs to the model in order to produce false results \cite{Biggio:ECML2013}. Usually, the examples used to fool a DL model at interface time are called adversarial examples, noticed for the first time by Szegedy et al. \cite{Szegedy:ICML2014}, are defined as follows:
\begin{equation}
    x^{*} = x + arg\min \{\|\delta\|:f(x+\delta)=t\}
    \label{eq:adversarial_samples}
\end{equation}
In other words, an adversarial example (or adversarial sample) $x^*$ is created by adding an imperceptible perturbation $\delta$ to the sample $x$ correctly classified. $\delta$ is computed by approximating the optimisation problem (defined in the Equation \ref{eq:adversarial_samples}) in an iterative way until the crafted adversarial sample is classified by the DL classifier $f(.)$ in targeted class $t$.
While attacks attempt to force the target DL models to misclassify using adversarial examples, defense techniques tend to strengthen the resilience of DL models against adversarial examples, while preserving the performance of DL models on legitimate instances (i.e. non adversarial samples). In fact, based on what kind of attack they contrast, Tariq et al. \cite{Tariq2020} classify the defense techniques as against evasion attacks, which can use detection of adversarial examples, adversarial training or defensive distillation. Adversarial training was introduced for the first time by Papernot et al. \cite{Papernot:CoRR2015}. Defensive distillation refers to a DL that produces a set of confidence levels for each training example. Whereas, defenses against poisoning attacks usually eliminate the extreme values that fall outside the relevant group as described in the framework proposed by Steinhardt et al. \cite{Steinhardt:NIPS2017}.



To evaluate the performance of defence methods, Qayyum et al. \cite{Qayyum:2020} list some principles such as defending against the adversary, testing the worst-case robustness and measuring the performance towards human-level abilities. In addition, they propose some common evaluation recommendations as well, including using both targeted and untargeted attacks, performing ablation (i.e. removing some defence components and testing again), diversifying test settings, evaluating defence on broader domains and analysing the transferability attacks. 

In this paper, we evaluate the methods for improving the robustness of DL models against adversarial attacks. In particular, DeepSec \cite{DeepSec} provides us with a unique platform to analyse the security of a DL model. Despite the fact that this platform already includes a series of attacks and defense techniques that are subsequently evaluated, we are interested in improving the robustness of DL models based on the recommendations in \cite{madry2018towards}. Extensive experimental results are presented to demonstrate the effectiveness of increasing the capacity of DL models to improve the robustness against adversarial attacks.

\section{Methodology}
\label{sec:DeepSec_problems}
The attacks and defenses in DeepSec \cite{DeepSec} are essentially PGD (Projected Gradient Descent) \cite{madry2018towards} and PAT (PGD Adversarial Training) \cite{madry2018towards}. However, the DeepSec implementation of PGD and PAT are different from \cite{madry2018towards} and, therefore, it loses some key aspects of the project.

In particular, \cite{madry2018towards} employ a ResNet model with wider layers in order to study the impact of the network capacity and they set $\epsilon=8$. Those are essential parts of the new approach they propose that are not considered in DeepSec \cite{DeepSec}.

Based on this observation, we propose to modify DeepSec accordingly to evaluate whether the optimisation problem and the relative solution proposed by \cite{madry2018towards} can improve the robustness of deep learning models against adversarial attacks. In addition, we focus on the CIFAR-10 \cite{CIFAR-10} dataset because existing methods have already achieved a state-of-the-art accuracy near 100\% on the MNIST \cite{MNIST} dataset.

At this point, we could have focused on developing a particular defence mechanism for a specific type of attack. However, we can take a different approach as in \cite{madry2018towards}, and create a unique mechanism to focus on both attacks and defense for any adversarial example. In the following sections, we will first review the basics of creating robust DL models.


\subsection{Creating robust Deep Learning models}
Madry et al. proposed a new approach in \cite{madry2018towards}, i.e. the robustness to adversarial attacks is seen under the perspective of a saddle point optimisation problem (or min-max). Min-max optimisation is a decision rule commonly used in AI for minimizing the possible loss for a worst-case (max loss) scenario.  
Not only does this technique allows for DL models robust to any class of attacks, but it provides us with a unique theoretical framework that includes both defenses and attacks.

In particular, Madry et al. \cite{madry2018towards} define the problem as follows:
\begin{equation}
    \min_{\theta} \rho(\theta), where~\rho(\theta) = \mathbb{E}_{(x,y)-\mathcal{D}} [ \max_{\delta \in s} L(\theta,x+\delta,y)]
    \label{eq:rho}
\end{equation}

In order to obtain this, the authors start by considering a standard classification task that employs a data distribution $\mathcal{D}$ over pairs of samples $x \in R^d$ and the corresponding labels $y \in [k]$. In addition, they assume that they are given a loss function, defined as $L(\theta,x,y)$, where $\theta \in R^p$ is the set of parameters for the model. The goal is to find the parameters $\theta$ that minimise the risk $\mathbb{E}_{(x,y)-\mathcal{D}} [ L(\theta,x,y)]$. Empiric risk minimisation (ERM) is a successful method to find the classifier with a small population risk.

Madry et al. \cite{madry2018towards} further define the attack model as: for each input $x$, a set of perturbations $S \subseteq R^d$ that formalise the manipulation power of the adversarial attack will be introduced. Recall the definition of adversarial examples from Equation \ref{eq:adversarial_samples} and we incorporate that in the population risk $\mathbb{E}_{\mathcal{D}}[L]$, we will obtain the saddle point problem described in Equation \ref{eq:rho}.

This robust optimisation problem can be viewed as an inner maximisation problem and an outer minimisation problem. The first one aims to create an adversarial version of a given data point x that achieves a high loss. Whereas, the second one’s goal is to find model parameters so that the adversarial loss given by the adversarial is minimised. As a result, when the parameters $\theta$ yield a (nearly) vanishing risk, the corresponding model is perfectly robust to attacks specified by our attack model. In other words, the unified view on attacks and defenses give an answer to both the question of how to produce effective adversarial examples and how to train a model free from adversarial samples. 

On the attack side, Madry et al. \cite{madry2018towards} focus on PGD (Projected Gradient Descent), which they claim to be a ‘universal’ adversary. While on the defence side, they use a training dataset augmented with adversarial samples created by PGD. In addition, they argue that if a network is trained to be robust against PGD adversaries, then it becomes robust against a wide range of other attacks as well. The formula can be seen below:
\begin{equation}
    x^{t+1} = \prod_{x+s} (x^t + \alpha~sgn(\nabla_x L(\theta,x,y))
\end{equation}

From here, we can see that PGD is a multi-step variant of FGSM (Fast Gradient Sign Method):
\begin{equation}
    x + \epsilon~sgn(\nabla_x L(\theta,x,y))
\end{equation}

Therefore, applying PGD solves the inner maximisation problem. As for the outer minimisation, i.e. finding model parameters that minimise the ‘adversarial loss’, this is done by applying SGD (Stochastic Gradient Descent) using the gradient of the loss at adversarial examples during training. These techniques employed in \cite{madry2018towards} result in successful optimisation of the saddle point problem and ultimately in the training of a robust DL model.

In other words, as we saw, the attack side of the problem is represented by the PGD, whereas on the defence side the authors employ the adversarial training technique called PAT (PGD Adversarial Training), which retrains the model on a dataset that includes the adversarial samples created by the PGD attack. Adversarial training is generally thought to be one of the most effective defence techniques, further contributing to the importance of the methods presented in \cite{madry2018towards}.


\section{Experimental results}
In the experiments, we evaluate the effectiveness of the DeepSec platform \cite{DeepSec}, different adversarial attacks (CW2 and PGD) and defenses (NAT and PAT) on benchmark datasets MNIST \cite{MNIST} and CIFAR10 \cite{CIFAR-10}.

\subsection{PGD attack and PAT defense} \label{sec:PGD_PAT}
For PGD and PAT, the experiments were run on CIFAR-10 \cite{CIFAR-10}. We started with the unmodified model, i.e. a ResNet, which reports an accuracy of 90.01\%. As a second step, we randomly select the clean candidate examples that will be used in the PGD attack, and 9001 samples were selected. Next, the parameters ($\epsilon = 0.1$ and $\epsilon_{iter} = 0.01$) suggested by the authors were used for creating the adversarial examples. As suggested by the authors. This returned a misclassification ratio (MR) of 100\%.

Using these adversarial examples generated, the PGD Adversarial Training (PAT) is executed with $\epsilon = 0.3137$, $number~of~steps = 7$ and $step~size = 0.007843$. This results in an accuracy of the retrained (i.e. trained on the dataset augmented with adversarial examples) model of 81.11\%.

Finally, the evaluations of the attack and the defence technique are reported in the PGD ($\epsilon=0.3137$) column in Table \ref{tab:tab3} and the PAT ($\epsilon=0.1$) column in Table \ref{tab:tab4}, respectively.

\subsection{The Wide ResNet Model}
As mentioned in Section \ref{sec:DeepSec_problems}, the DeepSec implementation of PGD and PAT lack some key aspect highlighted by Madry et al. \cite{madry2018towards}. In particular, Madry et al. \cite{madry2018towards} demonstrated the ResNet \cite{ResNet} model can be successfully trained against strong adversaries by 1) increasing the network capacity and 2) constructing adversarial examples with $\epsilon=8$.

Inspired by the recommendations, we included a new model, called {\it Wide ResNet}, by modifying the original ResNet as follows:
\begin{itemize}
    \item Changing the layer size: the values of layer 1, 2 and 3 and the Fully Connected (FC) are multiplied to a wide factor (10 in our experiments).
    \item Changing the $\epsilon$ from 0.1 (default in DeepSec) to 8 and the $\epsilon_{iter}$ to 0.875. The latter is due to the fact that Madry et al. \cite{madry2018towards} run the algorithm for 7 iterations, therefore $\epsilon$ is divided by the number of iterations: $8/7=0.875$.
    \item For the PAT, the new parameters are: $\epsilon=8$, $number~of~steps=10$ and the $step~size=2$ as suggested in \cite{madry2018towards}.
\end{itemize}

We follow the tests explained in Section \ref{sec:PGD_PAT} and the results are reported in the PGD ($\epsilon=8$) column in Table \ref{tab:tab3} and PAT ($\epsilon=8$) column in Table \ref{tab:tab4}.  The accuracy resulting obtained by the raw {\it Wide ResNet} model is 94.16\%, which is comparable to the 95.2\% obtained in \cite{madry2018towards}.

Then, after the candidate selection, we ran the PGD attack with the modified parameters, i.e. $\epsilon = 8$ and $\epsilon_{iter} = 0.875$. For the PAT test, the modified parameters are:$\epsilon = 8$, $number~of~steps = 10$ and $step~size = 2$, as in \cite{madry2018towards}. The final evaluation accuracy was 40.06\% as presented in the PAT ($\epsilon=8$) column in Table \ref{tab:tab4}.





\begin{table}[ht]
    \centering
    \begin{tabular}{p{1cm} p{1.5cm} p{1.5cm} p{1.5cm} p{1.5cm} p{2cm} }
         Metric & CW2 & CW2-DeepSec & PGD ($\epsilon=0.3137$) & PGD-DeepSec & PGD ($\epsilon=8$) \\
         \hline \hline
         MR & 99.8\% (ConvNet on MINST)  & 99.7\% (ConvNet on MINST)  & 100\% (ResNet on CIFAR10) & 100\% (ResNet on CIFAR10)  & 99.6\% (Wide ResNet on CIFAR10) \\ 
         \hline
         ACAC & 0.320 & 0.326 & 1.000 & 1.000 & 0.956\\ \hline
         ACTA & 0.312 & 0.318 & 0.000 & 0.000 & 0.002\\ \hline
         ALDp & & & & & \\
         L0 & 0.428 & 0.342 & 0.978 & 0.979 & 0.995\\
         L1 & 1.492 & 1.746 & 3.380 & 3.682 & 29.228\\
         L2 & 0.447 & 0.528 & 0.100 & 0.100 & 0.975\\ \hline
         ASS & 0.731 & 0.925 & 0.782 & 0.827 & 0.000\\ \hline
         PSD & 12.616 & 14.086 & 149.487 & 165.721 & 1331.845\\ \hline
         NTE & 0.001 & 0.003 & 1.000 & 1.000 & 0.978 \\ \hline
         RGB & 0.003 & 0.004 & 1.000 & 1.000 & 0.978 \\ \hline
    \end{tabular}
    \caption{Experimental results of the utility metrics of attacks}
    \label{tab:tab3}
\end{table}

\begin{table}[ht]
    \centering
    \begin{tabular}{p{1cm} p{1.5cm} p{1.5cm} p{1.5cm} p{1.5cm} p{2cm} }
         Metric & NAT & NAT-DeepSec & PAT ($\epsilon=0.1$) & PAT-DeepSec & PAT ($\epsilon=8$) \\
         \hline \hline
         ACC (Raw model) & 99.36\% (ConvNet on MINST) & 99.27\% (ConvNet on MINST)  & 90.01\% (ResNet on  CIFAR10) & 85.95\% (ResNet on  CIFAR10)  & 94.16\% (Wide ResNet on CIFAR10)\\ 
         \hline
         ACC (defence-enhanced model) & 99.27\% & 99.51\% & 81.11\% & 80.23\% & 40.25\%\\ \hline
         CAV & -0.09\% & 0.24\% & -8.90\% & -5.72\% & -49.76\%\\ \hline
         CRR & 0.25\% & 0.44\% & 3.92\% & 6.60\% &  2.61\%\\ \hline
         CSR & 0.34\% & 0.20\% & 12.82\% & 12.32\% & 52.37\%\\ \hline
         CCV & 0.26\% & 0.17\% & 16.58\% & 13.87\% & 31.58\%\\ \hline
         COS & 0.0009 & 0.0006 & 0.0683 & 0.0572 & 0.1345 \\ \hline
    \end{tabular}
    \caption{Experimental results of the utility metrics of defences}
    \label{tab:tab4}
\end{table}
\subsection{Experimental setup}
For running the experiments we used a PC equipped with an NVIDIA GeForce 1080 GPU and Ubuntu as the operating system.




\subsection{Evaluation Metrics}
We follow the literature to evaluate the performance of the adversarial attacks and defenses. Specifically, 9 metrics (Table \ref{tab:attack}) were used for evaluating the adversarial attacks as in \cite{DeepSec} and 5 metrics (Table \ref{tab:defence}) were used for evaluating the defence performance as in \cite{madry2018towards}.


\begin{table}[ht]
    \centering
    \begin{tabular}{|p{2.5cm}|p{8cm}|} \hline
        {\bf Misclassification Ratio (MR)} & the percentage of adversarial samples that are successfully misclassified\\ \hline 
        {\bf Average Confidence of Adversarial Class (ACAC)}  & the average prediction confidence towards the incorrect class\\ \hline
        {\bf Average Confidence of True Class (ACTC)} & evaluate to what extent the attacks escape from the ground truth, by averaging the prediction confidence of true classes for adversarial examples\\ \hline
         {\bf Average Lp Distortion (ALDp)} & the average normalized Lp (with $p=0,1,\infty$) norm distortion for all successful adversarial examples. The smaller ALDp is, the more imperceptible the adversarial examples are\\ \hline
        {\bf Average Structural Similarity (ASS)} & i.e. the average SSIM similarity \cite{Wang:TIP2014} (alternative to $Lp~norm$) between all successful adversarial examples and their original samples. Therefore, the greater ASS is, the more imperceptible the adversarial samples are\\ \hline
        {\bf Perturbation Sensitivity Distance (PSD)} \cite{Liu:TCSVT2010} & evaluates the human perception of perturbations. The smaller it is, the more imperceptible the adversarial examples are\\ \hline
         {\bf Noise Tolerance Estimation (NTE)} & calculates the gap between the probability of misclassified class and the maximum probability of all other classes. The higher NTE is, the more robust adversarial samples are \\ \hline
         {\bf Robustness to Gaussian Blur (RGB)} & measures how much a robust adversarial example maintain its misclassification effect after RGB is applied, which is used in pre-processing. The higher RGB is, the more robust adversarial examples are \\\hline
         {\bf Computation Cost (CC)} & the average runtime employed by the attacker to generate an adversarial example, i.e. the attack cost\\ \hline
    \end{tabular}
    \caption{Metrics for evaluating adversarial attacks}
    \label{tab:attack}
\end{table}

\begin{table}[ht]
    \centering
    \begin{tabular}{|p{2.5cm}|p{8cm}|} \hline
        {\bf Classification Accuracy Variance (CAV)} & since the most important metric for evaluating a DL model is its accuracy, and a defence technique should maintain this as much as possible on normal samples \\ \hline
        {\bf Classification Rectify/Sacrifice Ratio (CRR/CSR)} & CRR is defined as the percentage of testing samples that are misclassified by the model previously but correctly classified by the defence-enhanced model. Inversely, CSR is the percentage of testing samples that are correctly classified by the original model but misclassified by the defence-enhanced one. Therefore, CAV is equal to the difference between CRR and CSR \\ \hline
        {\bf Classification Confidence Variance (CCV)} & even though the accuracy might remain the same, sometimes the confidence of correctly classified samples can significantly decrease \\ \hline
        {\bf Classification Output Stability (COS)} & measure the classification output stability between the original model and the defence-enhanced model, we use JS divergence \cite{dagan-1997} to calculate the similarity of their output probability. We average the JS divergence between the output of original and defence-enhanced model on all correctly classified testing examples. \\ \hline
    \end{tabular}
    \caption{Metrics for evaluating defenses}
    \label{tab:defence}
\end{table}


\section{Results and Discussions}
The results of the experiments give us the evaluation utility metrics that are shown in Table \ref{tab:tab3} and \ref{tab:tab4}. While DeepSec \cite{DeepSec} focuses on comparing the metrics of different attacks and defences to determine which one is the best, we try to analyse the impact of the modification we brought, namely the wider ResNet and the higher value of $\epsilon$.

\subsection{Attack Utility Metrics}
In the first two columns in Table \ref{tab:tab3}, we reported the results obtained from the corresponding values presented in \cite{DeepSec}. This is done with the aim of determining to what extent the random component present in every DL algorithm will create a difference in results.

Despite not being exactly the same, the various utility metrics seem to be in a very similar range, therefore, not compromising their meaning. For instance, 
we can notice how the Carlini-Wagner (CW2) algorithm presents a very strong attacking ability, confirmed by the close to 100\% MR. In addition, the relatively low ACTC confirm a higher resilience of the CW2 attack to other models, such as a defence-enhanced one. Therefore, confirming its good attack ability.

CW2 also presents a good imperceptibility, as proven by the low value of ALD and ASS. However, it needs to be noticed the relatively high PSD, which is one of the best metrics to evaluate imperceptibility \cite{Liu:TCSVT2010}. This means that the attack is quite perceptible by humans.

Another disadvantage of CW2 seems to be its robustness, indeed it presents very low values in its robustness metrics (NTE, RGB and RIC).
Finally, the computational cost of CW2 is quite high. Despite not being the worst, CW2 is not the best for the amount of resources it needs. This similarity of results can also be noticed when comparing our second experiment on a PGD attack and the values reported in \cite{DeepSec}. In fact, both the experiments conducted 
find that PGD is one of the best attack methods available, with a state-of-the-art misclassification rate. In addition, PGD generated adversarial examples are way more robust to pre-processing than CW2. However, despite presenting a great imperceptibility (due to low ALD and ASS values), the PSD metric is extremely high, especially if compared to other attacks. This could result in an easy to detect adversarial examples for humans.

Finally, the Computational Cost is very low, especially when compared with other attacks. If we analyse the impact of increasing the dimension of the layers ({\it Wide ResNet}) and changing the value of $\epsilon$ from 0.3 to 8, we can notice a series of differences.
In general, the values seem to be just slightly worse, with an exception on the Average Structural Similarity (ASS) and on PSD. The first one seems to be more of a computational mistake rather than a plausible value, indeed such a low value is not reported by any type of attack. Whereas, the latter looks to be a worrying value which could make adversarial examples really easy to detect by a human operator.

\subsection{Defence utility metrics}
As for the defence utility metrics, we tend to obtain values that present a more noticeable difference from those reported in \cite{DeepSec}. In fact, while generally obtaining better accuracy on the raw models as reported in Table \ref{tab:tab4}, we notice a drop in accuracy on the defence-enhanced model that is more noticeable than the one obtained in \cite{DeepSec}. However, this is in line with the results obtained in \cite{madry2018towards}, i.e. a 47.1\% accuracy on the standard model and 50\% on the wide one. It, therefore, 
DeepSec \cite{DeepSec} has achieved better results.

Starting from the Naive Adversarial Training (NAT) defence technique, we notice how this algorithm increases the accuracy of the defence-enhanced model. In fact, according to the CRR value, 44\% of the samples misclassified by the raw model are classified correctly by the new model. While 20\% (CSR) of the samples are correctly classified by the raw model, but misclassified by the defence-enhanced one. Therefore, creating an increase in accuracy, as proven by the positive value of CAV (CAV=CRR-CSR).

However, this benefit brought by the NAT technique does not seem to be produced by the PGD Adversarial Training. In fact, in all the results relative to PAT we can notice a negative CAV, which represents a decrease in the accuracy of the new model in comparison to the original one.

It needs to be pointed out that the NAT technique was used on the MNIST dataset, which is simpler than the CIFAR10 one. 
As a result, most of the values that look less performing in Table \ref{tab:tab4} are to be ascribed to this higher complexity that generates a higher instability in the prediction of testing samples. In fact, other defence methods on MNIST have similar results to the NAT technique and PAT is among the best performing defence techniques on CIFAR10.

As for the change we brought to the model, we can notice a remarkable increase in the accuracy of the raw model (up to 94.16\%), due to a wider neural network. This gain in network capacity is essential to have more room to protect us against misclassification. However, the variation of epsilon seems to indicate a significant loss in the accuracy of the model to which the defence technique was applied.

Lastly, despite not being included in the defence utility metrics, the Computational Cost of PAT was a metric worth mentioning. In fact, the retraining on the adversarial augmented dataset required our machine for approximately 5 days or around 7000 minutes. This highlights a considerable weakness of this algorithm, i.e. its high demand for resources in order to obtain a robust model.

\section{Conclusions}
In this research, the results indicate that increasing $\epsilon$ to 8, as used by \cite{madry2018towards}, has generated worse metrics in the experiment condition set by DeepSec \cite{DeepSec}. However, operating on a wider ResNet has allowed us to gain more accuracy to be used to protect against misclassification caused by adversarial attacks and, therefore, create a more robust model.



Finally, as for evaluation, Carlini and Wagner \cite{Carlini:2017} propose an alternative method. In particular, they identify weaknesses in one of the most effective defence methods, i.e. defensive distillation. They develop three new attacks algorithms in order to be used as a benchmark for the effectiveness of a robust model and this is an interesting future direction to further evaluate the performance of adversarial attack and defense.

\bibliographystyle{unsrt}  
\bibliography{references}  

\end{document}